\documentclass[runningheads]{llncs}
\usepackage[english]{babel}
\usepackage{caption} 
\usepackage{graphicx}
\usepackage{amsmath}
\usepackage{algpseudocode}
\usepackage[boxed,vlined, linesnumbered]{algorithm2e}
\DontPrintSemicolon
\usepackage{import}
\usepackage{amssymb}
\usepackage{multirow}
\usepackage{bussproofs}
\usepackage{xcolor}
\usepackage{rotating}
\usepackage{tabularx}
\usepackage{cleveref}
\usepackage{todonotes}
\usepackage{makecell}
\usepackage{subcaption}
\usepackage{todonotes}

\begin{document}
\title{Guiding Word Equation Solving using Graph Neural Networks (Extended Technical Report)}

\newcommand{\answerTodo}[1]{\todo[color=green!40]{#1}} 
\newcommand*\samethanks[1][\value{footnote}]{\footnotemark[#1]}

\newcommand\SAT{\ensuremath{\mathrm{SAT}}\xspace}
\newcommand\UNSAT{\ensuremath{\mathrm{UNSAT}}\xspace}
\newcommand\UNKNOWN{\ensuremath{\mathrm{UNKNOWN}}\xspace}
\newcommand\BT{\ensuremath{\mathit{BT}}\xspace}
\newcommand\OURSOLVER{\textsf{DragonLi}\xspace}

\newcommand\myCall[2]{\ensuremath{\mathit{#1}(#2)}}

\SetKwRepeat{Do}{do}{while}
 \author{
   Parosh Aziz Abdulla
   \inst{1}
   \and 
   Mohamed Faouzi Atig  
    \inst{1}
 	\and
 	Julie Cailler 
 	\inst{2} 
 	\and
 	Chencheng Liang
 	\inst{1} 
    \and
 	Philipp R{\"u}mmer
 	\inst{1,2} 
 }
 \authorrunning{P. Abdulla et al.}
 %
 \institute{
 Uppsala University, Uppsala, Sweden\\
 \email{parosh.abdulla@it.uu.se}\\
 \email{mohamed\_faouzi.atig@it.uu.se}\\
 \email{chencheng.liang@it.uu.se}\\
 \and
 University of Regensburg, Regensburg, Germany\\
 \email{philipp.ruemmer@ur.de}\\
 \email{julie.cailler@ur.de}\\
 }

\maketitle              

\begin{abstract}

This paper proposes a Graph Neural Network-guided algorithm for solving word equations, based on the well-known Nielsen transformation for splitting equations.
The algorithm iteratively rewrites the first terms of each side of an equation, giving rise to a tree-like search space.
The choice of path at each split point of the tree significantly impacts solving time, motivating the use of Graph Neural Networks (GNNs) for efficient split decision-making.
Split decisions are encoded as multi-classification tasks, and five graph representations of word equations are introduced to encode their structural information for GNNs.
The algorithm is implemented as a solver named \OURSOLVER.
Experiments are conducted  on artificial and real-world benchmarks. 
The algorithm performs particularly well on satisfiable problems. 
For single word \mbox{equations}, \mbox{\OURSOLVER} can solve significantly more problems than well-established string solvers.
For the conjunction of multiple word equations, \OURSOLVER is competitive with state-of-the-art string solvers.

\end{abstract}

\section{Introduction}

Over the past few years, reasoning within specific theories, including arithmetic, arrays, or algebraic data structures, has become one the main challenges in automated reasoning. 
To address the needs of modern applications, new techniques have been developed, giving rise to \emph{SMT} (Satisfiability Modulo Theories) solvers. 
SMT solvers implement efficient decision procedures and reasoning methods for a wide range of theories, and are used in applications such as verification.

Among the theories supported by SMT solvers, the theory of \emph{strings} has in particular received  attention in the last years. Strings represent one of the most important data-types in programming, and string constraints are therefore relevant in various domains, from text processing to database management systems and web applications.
One of the simplest kind of constraints supported by the SMT-LIB theory of strings~\cite{BarFT-RR-17} are \emph{word equations,} i.e., equations in a free semigroup~\cite{Khm71}.
Makanin's work~\cite{Mak77} demonstrated the decidability of the word equation problem, which was later confirmed to be in PSPACE~\cite{10.1145/1132516.1132584}.
However, even the leading SMT solvers with support for string constraints (including \textsf{cvc5}~\cite{10.1007/978-3-030-99524-9_24}, \textsf{Z3}~\cite{demoura2008z}, \textsf{Norn}~\cite{10.1007/978-3-319-21690-4_29}, \textsf{TRAU}~\cite{8602997}, \textsf{Ostrich}~\cite{DBLP:journals/pacmpl/ChenHLRW19}, \textsf{Woorpje}~\cite{day2019solving}, and \textsf{Z3-Noodler}~\cite{10.1007/978-3-031-57246-3_2}) tend to be incomplete for proving the unsatisfiability of word equations, illustrating the hardness of the theory.

Solving a word equation is to check for the existence of string values for variables that make equal both sides of the equation.
For example, consider the equation $Xab=YaZ$, where
$X, Y$, and $Z$ are variables ranging over strings, and $a$ and $b$ are letters.
This equation is satisfiable and has multiple solutions. For example, assigning $a$ to both $X$ and $Y$ and $b$ to $Z$ results in $aab=aab$. 

This paper presents an algorithm that makes use of \emph{Graph Neural Networks} (GNN)~\cite{DBLP:journals/corr/abs-1806-01261} in order to solve word equations.
It is an extension of the method proposed in~\cite{10.1007/978-3-319-08867-9_10} and implemented in \textsf{Norn}~\cite{10.1007/978-3-319-21690-4_29}, referred to as the \emph{split algorithm.} The split algorithm is, in turn, based on the well-known Nielsen transformation~\cite{Nielsen1917}.
It builds a proof tree by iteratively applying a set of inference (split) rules on a word equation.

One critical aspect of the algorithm lies in selecting the next branch to be explored while constructing the proof tree, which significantly influences the solving time.
To address this, we present a heuristic that leverages deep learning to determine the exploration order of branches.
GNNs~\cite{DBLP:journals/corr/abs-1806-01261} represent one of the paradigms in neural network research, tailored for non-Euclidean, graph-structured data. This makes them suited for scenarios where data points are interconnected, such as social networks~\cite{10.1145/3308558.3313488}, molecular structures~\cite{10.5555/3305381.3305512}, programs~\cite{10.5555/3015812.3016002,DBLP:journals/corr/abs-1711-00740}, and logical formulae~\cite{10.5555/3294996.3295038,10.1007/978-3-030-51054-1_29,DBLP:journals/corr/abs-1905-10006,LPAR2023:Guiding_an_Instantiation_Prover}.
  Our work represents, to the best of our knowledge, the first use of deep learning in the context of word equations.

\begin{figure}[t]
  \begin{center}
    \includegraphics[width=0.9\textwidth,trim=20 5 20 5]{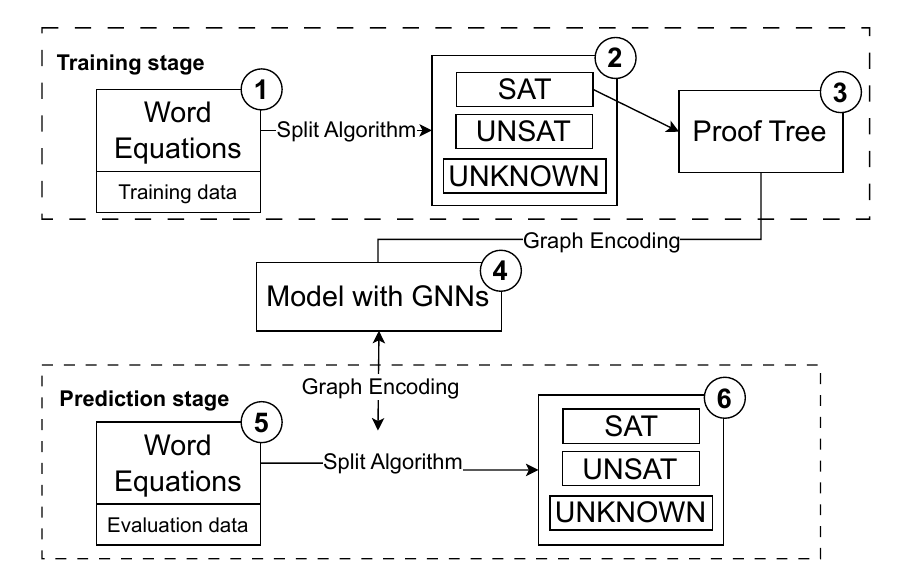}
  \end{center}
      
	\caption{The workflow diagram for the training and prediction stage}
	\label{fig:workflow-diagram}
\end{figure}

Figure~\ref{fig:workflow-diagram} illustrates the workflow of our approach.
During the training stage, we initially employ the split algorithm (without GNN guidance) to solve word equation problems drawn from a training dataset. For each satisfiable (SAT) problem, we generate a corresponding proof tree. Within this tree, each branch (pair of a node and a direct child) is evaluated to determine whether it leads to a solution, as well as its distance to the corresponding leaf.
Based on this information, we assign labels to each branch to indicate whether it is a favorable choice for reaching a solution. Subsequently, we encode each branching point comprising the node and its child nodes as a graph.
These graph representations, along with their associated labels, are then used to train the GNN model.

In the prediction stage, word equations from an evaluation dataset are processed using the split algorithm, now guided by the GNN model.
At each branching point, the current branch is first transformed into a graph representation, which is in turn fed to the trained deep learning model with GNNs. The trained model, using its learned insights, advises on which branch should be prioritized and explored first.

We implemented this algorithm in the \OURSOLVER tool.
Experiments were conducted using four word equation benchmarks: three of them are artificially generated and inspired by \textsf{Woorpje}~\cite{day2019solving}; a fourth one is extracted from SMT-LIB benchmarks~\cite{smtlib:benchmark} and encodes real-world problems. 
Results show that for SAT problems, the pure split algorithm without GNNs is already competitive with some leading string solvers (\textsf{Z3}~\cite{demoura2008z}, \textsf{cvc5}~\cite{10.1007/978-3-030-99524-9_24}, \textsf{Ostrich}~\cite{DBLP:journals/pacmpl/ChenHLRW19}, \textsf{Woorpje}~\cite{day2019solving}, \textsf{Z3-Noodler}~\cite{10.1007/978-3-031-57246-3_2}), while it performs less well on UNSAT problems.
We conjecture that this is due to the (relatively straightforward) depth-first search performed by our implementation of the split algorithm, which is a good strategy for finding solutions, whereas other solvers devote more time (e.g., using length reasoning) to show that formulas are unsatisfiable.
Enabling GNN guidance in \OURSOLVER uniformly improves performance on SAT problems, allowing it to outperform all other solvers in one specific benchmark. Specifically, in Benchmark~2, the GNN-guided version of \OURSOLVER solves 115\% more SAT problems than its non-GNN-guided counterpart and 43.0\% more than the next best string solver, \textsf{Woorpje}.


In summary, the contributions of this paper are as follows:
\begin{itemize}
\item We define a proof system based on the split algorithm for
  solving word equations, tailored to combining symbolic reasoning
  with GNN-based guidance.
  %
\item Based on the proof system, we introduce an algorithm for
  integrating GNN-based guidance with the proof system.
\item To train a GNN on the data obtained from solving word equations,
  we present five possible graph representations of word equations.
\item We present an extensive experimental evaluation on four word
  equation benchmarks, comparing, in particular, the different graph
  encodings and different backtracking strategies of the algorithm.
\end{itemize}


\section{Preliminaries}\label{section:preliminaries}
We start by defining the syntax of word equations, as well as the notion of satisfiability. Then, we explain the fundamental mechanism of Graph Neural Networks (GNNs), along with a description of the specific GNN model we have employed in our experiments.

\subsubsection{Word Equations.}
\label{section:preliminaries:word-equations}

We assume a finite non-empty alphabet $\Sigma$ and write
$\Sigma^{*}$ for the set of all strings (or words) over $\Sigma$.
We work with a set $\Gamma$ of string variables, ranging over words in $\Sigma^{*}$, and denote the empty string by $\epsilon$. The symbol $\cdot$ denotes the concatenation of two strings; in our examples, we often write~$uv$ as shorthand for $u \cdot v$.
The syntax of word equations is defined as follows:
\begin{align*}
  \text{Formulae} ~ \phi & ::= \mathit{true} \mid e \land \phi &
 \text{Words} ~ w & ::= \epsilon \mid t \cdot w   \\
 \text{Equations} ~ e & ::= w = w   &
          \text{Terms} ~ t & ::= X \mid c  
\end{align*}
where $X \in \Gamma$ ranges over variables and $c \in \Sigma$ over letters.

\begin{definition}[Satisfiability of word equations]
  A formula~$\phi$ is \emph{satisfiable} if there exists a substitution~$\pi: \Gamma \rightarrow \Sigma^{*}$ such that, when each variable $X\in \Gamma $ in $\phi$ is replaced by $\pi(X)$, all equations in $\phi$ are satisfied.
\end{definition}

\subsubsection{Graph Neural Networks.}



A \emph{Graph Neural Network} (GNN)~\cite{DBLP:journals/corr/abs-1806-01261} uses \emph{Multi-Layer Perceptrons} (MLPs) to extract features from a given graph.
MLPs, also known as multi-layer neural networks~\cite{GoodBengCour16}, transform an input space to make different classes of data linearly separable, and this way learn representations of data with multiple levels of abstraction.
Each layer of an MLP consists of neurons that apply a nonlinear transformation to the inputs received from the previous layer.
This allows MLPs to learn increasingly complex patterns as data moves from the input layer to the output layer.

\emph{Message passing-based GNNs} (MP-GNNs) \cite{DBLP:journals/corr/GilmerSRVD17} are designed to learn features of graph nodes (and potentially the entire graph) by iteratively aggregating and transforming feature information from the neighborhood of a node.
For instance, if we represent variables in a word equation by nodes in a graph, then node features could represent symbol type (i.e., being a variable), possible assignments, or the position in the word equation.

Consider a graph $ G = (V, E) $, with $ V $ as the set of nodes and $ E \subseteq V \times V $ as the set of edges. Each node $ v \in V $ has an initial representation $ x_{v} \in \mathbb{R}^{n}$ and a set of neighbors $ N_{v} \subseteq V $.
In an MP-GNN comprising $ T $ message-passing steps, node representations are iteratively updated. At each step $ t $, the representation of node $ v $, denoted as $ h_{v}^{t} $, is updated using the equation:
\begin{equation}\label{eq:MPGNN}
h_{v}^{t} = \eta_{t}(\rho_{t}(\{h_{u}^{t-1} \mid u \in N_{v}\}), h_{v}^{t-1}),
\end{equation}
where $ h_{v}^{t} \in \mathbb{R}^{n} $ is the updated representation of node $ v $ after $ t $ iterations, starting from the initial representation $ h_{v}^{0} = x_{v} $.
The node representation of $u$ in the previous iteration $t-1$ is $h_{u}^{t-1}$, and node $u$ is a neighbor of node $v$.
In this context, $\rho_{t}:(\mathbb{R}^{n})^{|N_{v}|} \rightarrow \mathbb{R}^{n}$ is an aggregation function with trainable parameters (e.g., an MLP followed by sum, mean, min, or max) that aggregates the node representations of $v$'s neighboring nodes at the $t$-th iteration.
Along with this, $\eta_{t}:(\mathbb{R}^{n})^2\rightarrow \mathbb{R}^{n}$ is an update function with trainable parameters (e.g., an MLP) that takes the aggregated node representation from $\rho_{t}$ and the node representation of $v$ in the previous iteration as input, and outputs the updated node representation of $v$ at the $t$-th iteration.

MP-GNNs operate under the assumption that node features can capture structural information from  long-distance neighbors by aggregating and updating features of neighboring nodes.
After $T$ message-passing steps, an MP-GNN yields an updated node representation (feature) that includes information from neighbors within a distance of $T$, applicable to various downstream tasks like node or graph classification.

In this study, we choose  \emph{Graph Convolutional Networks} (GCNs)~\cite{kipf2017semisupervised} to guide our algorithm. 
In GCNs, the node representation $h_{v}^{t}$ of $v$ at step $t \in \{1, ..., T\}$ where $T \in \mathbb{N}$ is computed by 
\begin{equation}\label{eq:GCN}
 h_{v}^{t} = \text{ReLU}(\text{MLP}^{t}(\text{mean}\{h_{u}^{t-1} \mid u \in N_{v} \cup \{v\}\})),
\end{equation}
where each $\text{MLP}^{t}$ is a fully connected neural network,  ReLU (Rectified Linear Unit)~\cite{agarap2018deep} is the non-linear function $f(x)=max(0,x)$, and $h_{v}^{0}=x_{v}$.



\section{Search Procedure and Split Algorithm}

In this section, we define our proof system for word equations,  the notion of a proof tree, and show soundness and completeness.
We then introduce an algorithm to solve a conjunction of word equations.

\subsection{Split Rules}
\label{section:split-rules}

We introduce four types of proof rules in Figure~\ref{fig:split_rules}, each corresponding to a specific situation. The proof rules are inspired by~\cite{10.1007/978-3-319-08867-9_10}, but streamlined and formulated differently.
Each proof rule is of the form:
\begin{center}
\def\arraystretch{2}
	\begin{tabularx}{8cm}{lX|X|X}
		\multirow{2}{*}{$\mathit{Name}$} & \multicolumn{3}{c}{$P$} \\ \cline{2-4}
							   & \makecell{$[\mathit{cond_1}]$\\$C_1$} & \makecell{\dots} & \makecell{$[\mathit{cond_n}]$\\$C_n$} 
	\end{tabularx}
\end{center}
Here, $\mathit{Name}$ is the name of the rule, $P$ is the premise, and $C_i$s are the conclusions.
Each $\mathit{cond_i}$ is a substitution that is applied implicitly to the corresponding conclusion~$C_i$, describing the case handled by that particular branch.
In our case, $P$ is a conjunction of word equations and each $C_i$ is either a conjunction of word equations or a final state, \SAT or \UNSAT.


To introduce our proof rules, we use distinct letters~$a, b \in \Sigma$ and variables $X, Y \in \Gamma$, while $u$ and $v$ denote sequences of letters and variables.

Rules $R_{1}$, $R_{2}, R_{3}$, and $R_{4}$ (Figure~\ref{fig:rules_CNF}) define how to simplify word equations, and how to handle equations in which one side is empty. In $R_3$, note that the substitution~$X \mapsto \epsilon$ is applied to the conclusion~$\phi$.
Rules $R_{5}$ and $R_{6}$ (Figure~\ref{fig:rules_Rcc}) refer to cases in which each word starts with a letter. The rules simplify the current equation, either by removing the first letter, if it is identical on both sides ($R_{5}$), or by concluding that the equation is \UNSAT ($R_{6}$).
Rule $R_{7}$ (Figure~\ref{fig:rules_Rvc}) manages cases where one side begins with a letter and the other one with a variable. The rule introduces two branches, since the variable must either denote the empty string~$\epsilon$, or its value must start with the same letter as the right-hand side.
Rule $R_8$ and $R_{9}$ (Figure~\ref{fig:rules_Rvv}) handle the cases in which both sides of an equation start with a variable, implying that either both variables have the same value or the value of one is included in the value of the other.

We implicitly assume symmetric versions of the rules $R_{3}$, $R_{4}$, and $R_{7}$, swapping left-hand side and right-hand side of the equation that is rewritten. For instance, the symmetric rule for $R_{3}$ would have premise $\epsilon = X \wedge \phi$.

\begin{figure}[t]
	\fbox{\parbox{0.975\linewidth}{
            \vspace*{-0.2cm}
		\begin{minipage}{\textwidth}
			\begin{minipage}{0.2\linewidth}
				\begin{prooftree}
					\rootAtTop
					\AxiomC{\SAT}
					\LeftLabel{$R_{1}$}
					\UnaryInfC{$\mathit{true}$}
				\end{prooftree}
			\end{minipage}%
			\begin{minipage}{0.25\linewidth}
				\begin{prooftree}
					\rootAtTop
					\AxiomC{$\phi$}
					\LeftLabel{$R_{2}$}
					\UnaryInfC{$\epsilon=\epsilon \wedge \phi$}
				\end{prooftree}
			\end{minipage}%
			\begin{minipage}{0.25\linewidth}
				\vspace{8pt}
				\begin{prooftree}
					\rootAtTop
					\AxiomC{$\phi$}
					\noLine
					\UnaryInfC{$[X \mapsto \epsilon]$}
					\LeftLabel{$R_{3}$}
					\UnaryInfC{$X=\epsilon \wedge \phi$}
				\end{prooftree}
			\end{minipage}%
			\begin{minipage}{0.25\linewidth}
				\begin{prooftree}
					\rootAtTop
					\AxiomC{\UNSAT}
					\LeftLabel{$R_{4}$}
					\UnaryInfC{$a \cdot u=\epsilon \wedge \phi$}
				\end{prooftree}
			\end{minipage}%
			\vspace{0.2cm}
			\centering
			with $X\in \Gamma$ and $a \in \Sigma$.
			\subcaption{Simplification rules}
			\label{fig:rules_CNF}
		\end{minipage}

		\vspace{0.2cm}
			
			\begin{minipage}{\textwidth}

				\begin{minipage}{0.50\linewidth}
					\begin{prooftree}
						\rootAtTop
						\AxiomC{\shortstack{$u = v \land \phi$}}
						\LeftLabel{$R_{5}$}
						\UnaryInfC{$a \cdot u = a \cdot v \land \phi$}
					\end{prooftree}
				\end{minipage}%
				\begin{minipage}{0.50\linewidth}
					\begin{prooftree}
						\rootAtTop
						\AxiomC{\shortstack{\UNSAT}}
						\LeftLabel{$R_{6}$}
						\UnaryInfC{$a \cdot u = b \cdot v \land \phi$}
					\end{prooftree}
				\end{minipage}%
				\vspace{0.2cm}
				\centering
				with $a, b$ two different letters from $\Sigma$.
				\subcaption{Letter-letter rules}
				\label{fig:rules_Rcc}
			\end{minipage}

			\vspace{0.5cm}

			\begin{minipage}{\textwidth}
				\begin{minipage}{\linewidth}
					\centering
					\begin{tabularx}{6cm}{lX|X}
						\multirow{2}{*}{$R_{7}$} & \multicolumn{2}{c}{$X \cdot u = a \cdot v \land \phi$} \\ \cline{2-3}
						                       & \makecell{$[X \mapsto \epsilon]$                         \\$u = a \cdot v \land \phi$} & \makecell{$[X \mapsto a \cdot X']$\\$X'\cdot u = v \land \phi$}
					\end{tabularx}
				\end{minipage}\hfill
				\vspace{0.2cm}
				\centering
				with $X'$ a \emph{fresh} element of $\Gamma$.
				\subcaption{Variable-letter rules}
				\label{fig:rules_Rvc}
			\end{minipage}

			\vspace{0.5cm}

			\begin{minipage}{\textwidth}
				\begin{minipage}{\textwidth}
					\centering
						\begin{tabularx}{8cm}{lX|X|X}
							\multirow{2}{*}{$R_{8}$} & \multicolumn{3}{c}{$X \cdot u = Y \cdot v \land \phi$} \\ \cline{2-4}
							& \makecell{$[X \mapsto Y]$\\$u = v \land \phi$}  & \makecell{$[X \mapsto Y \cdot Y']$\\$Y'\cdot u = v \land \phi$} & \makecell{$[Y \mapsto X \cdot X']$\\$ u = X' \cdot v \land \phi$} 
						\end{tabularx}
				\end{minipage}
			
				\vspace{0.2cm}
				\centering
				with $X\neq Y$ and $X', Y'$ \emph{fresh} elements of $\Gamma$.
				
				\vspace{0.2cm}
					\begin{minipage}{\linewidth}
						\centering
						\begin{tabularx}{3cm}{lX}
							\multirow{2}{*}{$R_{9}$} & $X \cdot u = X \cdot v \land \phi$ \\ \cline{2-2}
							& \makecell{$u =  v \land \phi$ \\ $~$} 
						\end{tabularx}
					\end{minipage}
						\vspace{-0.5cm}
				\subcaption{Variable-variable rules}
				\label{fig:rules_Rvv}
			\end{minipage}

		}}
	\caption{Rules of the proof system for word equations}
	\label{fig:split_rules}
\end{figure}

Although our proof system is not complete for proving the unsatisfiability of word equations, we can observe that the proof rules are sound and locally complete.
A proof rule is said to be \emph{sound} if the satisfiability of the premise implies the satisfiability of one of the conclusions. It is said to be \emph{locally complete} if the satisfiability of one of the conclusions implies the satisfiability of the premise.
%

\begin{lemma}\label{lemma:1}
	The proof rules in Figure~\ref{fig:split_rules} are sound and locally complete.
\end{lemma}


\subsection{Proof Trees}

Iteratively applying the proof rules to a conjunction of word equations gives rise to a \emph{proof tree} growing downwards. Given the proof rules $R_{1},\ldots,R_{9}$  in Figure~\ref{fig:split_rules}, we represent a proof tree as a tuple $\tau = (N,\alpha,E,\lambda)$ where: 
\begin{itemize}
\item $N$ is a finite set of nodes;
\item $E \subseteq N \times N$ is a set of edges, such that $(N, E)$ is a directed tree. An edge $(n_{i},n_{j})\in E$ implies that $n_{j}$ is derived from $n_{i}$ by applying a proof rule;
\item $\alpha : N \to \mathit{For} \cup \{\text{\SAT}, \text{\UNSAT}\}$ is a function mapping each node~$n \in N$ to a formula or to a label \SAT, \UNSAT;
\item $\lambda : E\rightarrow R$ is a function that assigns to each edge a proof rule.
\end{itemize}
A \emph{path} in the proof tree is a sequence of edges starting from the root and ending with a leaf node.
Due to local completeness, if there is a leaf node that is \SAT, then the word equation at the root node is satisfiable. Due to soundness, if all the leaf nodes are \UNSAT, then the formula at the root node is unsatisfiable. 

\begin{figure}[t]
  \begin{center}
    \fbox{\includegraphics[width=\textwidth-2\fboxsep,trim=50 10 20 5]{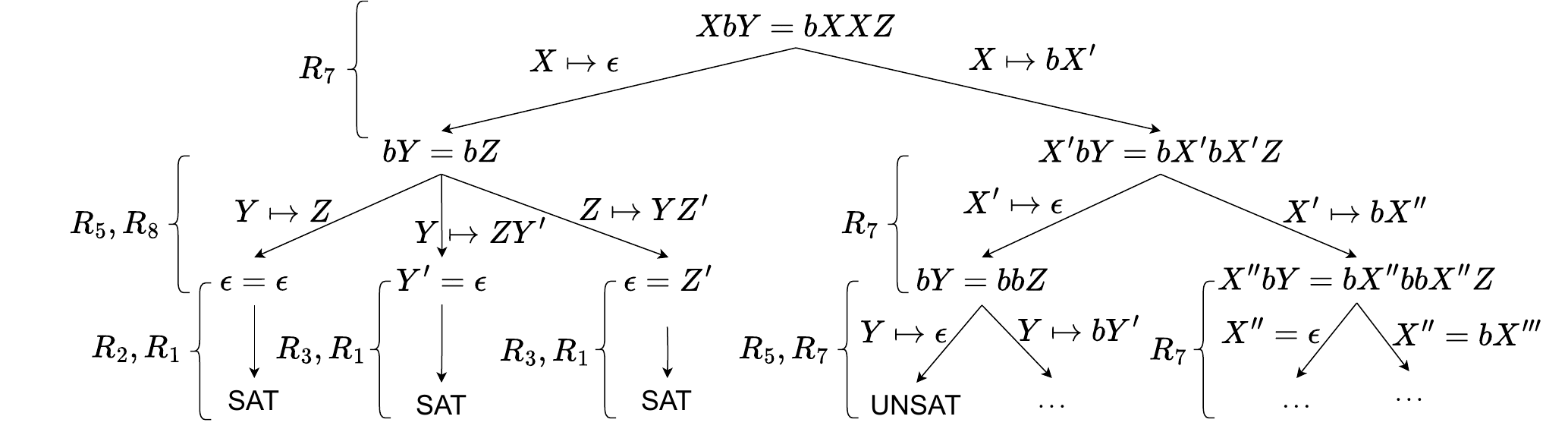}}
  \end{center}

  \caption{Proof tree resulting from the word equation $XbY = bXXZ$}
  \label{figure:proof-tree}
\end{figure}

Figure~\ref{figure:proof-tree} illustrates the proof tree generated by applying the proof rules in Figure~\ref{fig:split_rules} on the word equation $\phi = (XbY = bXXZ)$.
In this example, $b\in \Sigma$ and $X,Y,Z \in \Gamma$.
The application of $R_{7}$ on the root generates two branches. While exploring the left branch first yields a solution (SAT) at a depth of 3, iteratively navigating through the right branch of $R_{7}$ leads to non-termination since the length of the word equation keeps increasing. 

\subsection{GNN-Guided Split Algorithm}
\label{section:GNN-Guided_Split_Algorithm}
We use the proof rules in Figure~\ref{fig:split_rules} and the idea of iterative deepening from~\cite{korf1985IDFS} (combination of depth- and breadth-first search in a tree) to solve word equations, as shown in Algorithm~\ref{algorithm:SolveEqs}.
This algorithm aims to check the satisfiability of word equations~$\phi = (\bigwedge_{i=1}^n w_{i}^{l} = w_{i}^{r})$.

Algorithm~\ref{algorithm:SolveEqs} receives as parameter a backtrack strategy~$\BT \in \{\BT_1, \BT_2,$ $\BT_3\}$, which determines when to stop exploring a path of the proof tree and return to a previous branching point.
The algorithm calls the function $\mathit{solveEqsRec}$ (Algorithm~\ref{algorithm:SolveEqsRec}), which returns the satisfiability status by exploring a proof tree recursively.
At each branching point in this tree, if at least one child node is \SAT, the algorithm concludes that $\phi$ is \SAT and terminates (Line~\ref{alg:sat} of $\mathit{solveEqsRec}$).
Conversely, if every child node is \UNSAT, the current node is marked as \UNSAT (Line \ref{alg:unsat} of $\mathit{solveEqsRec}$), and the algorithm backtracks to the last branching point.
A formula~$\phi$ is considered \UNSAT only after all branches have been checked and found to be \UNSAT.

\begin{figure}

\begin{algorithm}[H]
	\caption{The top-level algorithm $\mathit{solveEqs}$ for word equations}
	\label{algorithm:SolveEqs}
	\SetAlgorithmName{Algorithm}{}{}
	\KwIn{
          \begin{tabular}[t]{@{}l@{}}
            Alphabet~$\Sigma$ and variables~$\Gamma$;\\
            Word equations $\phi = (\bigwedge_{i=1}^n w_{i}^{l} = w_{i}^{r})$;\\
            Backtrack strategy $\BT \in \{\BT_1,\BT_2, \BT_3\}$;\\
            Global backtrack limits $l_{\BT_2}, l_{\BT_2}^{step}, l_{\BT_3}$.
          \end{tabular}
	}
	\KwOut{The status of $\phi$: \SAT, \UNSAT, or \UNKNOWN}
	
	\Begin{

           $\mathit{res} \leftarrow \UNKNOWN$\;

		\If{$\BT \in \{\BT_1, \BT_2\}$}{
			$\mathit{res} \leftarrow \myCall{solveEqsRec}{\phi, 0, \BT, \Sigma, \Gamma}$\; \label{alg:bt1}		
		}
		\If{$\BT= \BT_3$}{ 	
			\Do{$\mathit{res} \neq \UNKNOWN$}{ \label{alg:bt3_start}			
				$\mathit{res} \leftarrow \myCall{solveEqsRec}{\phi, 0, \BT_{3},\Sigma,\Gamma}$\;
				$l_{\BT_{3}} \leftarrow l_{\BT_{3}}+1$\;	\label{alg:bt3_end}
			}
		}
		\Return $\mathit{res}$\;
	}
\end{algorithm}

\bigskip

\begin{algorithm}[H]
	\caption{Recursive exploration $\mathit{solveEqsRec}$ of word equations}
	\label{algorithm:SolveEqsRec}
	\SetAlgorithmName{Algorithm}{}{}
	\KwIn{
          \begin{tabular}[t]{@{}l@{}}
            Alphabet~$\Sigma$ and variables~$\Gamma$;\\
            Word equations $\phi = (\bigwedge_{i=1}^n w_{i}^{l} = w_{i}^{r})$;\\
            Backtrack strategy $\BT \in \{\BT_1,\BT_2, \BT_3\}$;\\
            Current exploration depth $\mathit{currentDepth}$;\\
            Global backtrack limits $l_{\BT_2}, l_{\BT_2}^{step}, l_{\BT_3}$.
          \end{tabular}
	}
	\KwOut{The state of $\phi$: \SAT, \UNSAT, or \UNKNOWN}

	\Begin{

	\If{$\BT=\BT_2 \wedge \mathit{currentDepth} \geq l_{\BT_{2}}$}{
		$l_{\BT_2} \leftarrow l_{\BT_2}+l_{\BT_2}^{step}$\;\label{alg:bt2_rec_start}
		\Return{\UNKNOWN}\;\label{alg:bt2_rec_end}
	}	
	\If{$\BT=\BT_3 \wedge \mathit{currentDepth} \geq l_{\BT_{3}}$}{
		\Return{\UNKNOWN}\;\label{alg:bt3_rec}
	}

	$\mathit{branches} \leftarrow \myCall{applyRules}{\phi,\Sigma,\Gamma}$\;
	
	$\mathit{branches} \leftarrow \myCall{orderBranches}{\mathit{branches}}$\;
	\label{alg:orderBranches}
	
	$\mathit{unknownFlag} \leftarrow \mathit{false}$\;
	
	\ForEach{$\mathit{child} \text{~in~} \mathit{branches}$}{
		$\mathit{res} \leftarrow \myCall{solveEqsRec}{\mathit{child}, \mathit{currentDepth}+1, \BT, \Sigma, \Gamma}$\;
		
		\If{$\mathit{res} = \SAT$}{
			\Return \SAT\; \label{alg:sat}
		}
		\If{res = \UNKNOWN}{
                  $\mathit{unknownFlag} \leftarrow \mathit{true}$\;
		}
		
	}
	\eIf{$\mathit{unknownFlag}$}{\Return \UNKNOWN\;}{\Return \UNSAT\; \label{alg:unsat}}
}
\end{algorithm}

\end{figure}

We explore three different backtrack strategies, $\BT_1$, $\BT_2$, and $\BT_3$:
\begin{itemize}
	\item $\BT_1$: This strategy performs depth-first search until it finds a \SAT node or exhausts all branches to conclude \UNSAT. It may lead to non-termination of the algorithm in proof trees with infinite branches, and can miss solutions; an example for this is  the rightmost branch of Figure~\ref{figure:proof-tree}.
	\item $\BT_2$: This hybrid strategy imposes a limit, $l_{\BT_{2}}$, on the depth to which a proof branch is explored.
	When the maximum depth $l_{\BT_{2}}$ is reached, the proof search backtracks to the last branching point, and $l_{\BT_{2}}$ is globally increased by $l_{\BT_{2}}^{step}$ (line~\ref{alg:bt2_rec_start} of $\mathit{solveEqsRec}$). Similarly as $\BT_1$, this strategy  can miss solutions of word equations.
	\item $\BT_3$: This strategy performs the classical depth-first search with iterative deepening, by setting an initial limit $l_{\BT_{3}}$ on the exploration depth. This limit is increased (line~\ref{alg:bt3_end} of $\mathit{solveEqs}$) when no node with label \SAT is found but the tree was not fully explored yet. This strategy is complete in the sense that it will eventually find a solution for every satisfiable formula.
\end{itemize}

The performance and termination of the algorithm are highly influenced by the order in which we explore the proof tree. This order is determined by the $\mathit{orderBranches}$ function (Line~\ref{alg:orderBranches} of $\mathit{solveEqsRec}$).
Our main goal in this paper is to study whether the integration of GNN models within $\mathit{orderBranches}$ is able to optimise solving time or make it more likely for the algorithm to terminate.

For a conjunction of multiple word equations, deciding which word equation to work on first is also important for performance.
Our current proof rules only rewrite the leftmost equation in a conjunction; reordering word equations is beyond the scope of this paper. We discuss this point further in Section~\ref{section:conclusion-and-future-work}.

The correctness of Algorithm~\ref{algorithm:SolveEqs} directly follows from the soundness and local completeness of the proof rules in Figure~\ref{fig:split_rules}:

\begin{lemma}[Soundness of Algorithm~\ref{algorithm:SolveEqs}]
	For a conjunction of word equations $\phi$, if Algorithm~\ref{algorithm:SolveEqs} terminates with the result \SAT or \UNSAT, then $\phi$ is \SAT or \UNSAT, respectively. 
\end{lemma}

%
%
\section{Guiding the Split Algorithm}
\label{section:learning}
This section describes how to train and apply the GNNs in the \emph{orderBranches} function in Algorithm~\ref{algorithm:SolveEqsRec}. 
We start by describing five graph representations for a conjunction of word equations, which encode word equations in form of text to graph representations to be readable by GNNs. Then, we explain how to  train our classification tasks on GNNs and collect the training data. Finally, we describe different ways to apply the predicted results back to algorithm.
 
\subsection{Representing Word Equations by Graphs}
\label{subsection:graph-representation}
Graph representations can capture the structural information in word equations and are the standard input format for GNNs. To understand the impact of the graph structure on our framework, we have designed five graph representations for word equations.

In order to extract a single graph from the equations, we first translate a conjunction~$\bigwedge_{i=1}^n w_{i}^{l} = w_{i}^{r}$ of word equations to a single word equation, by inserting a distinguished letter $\# \notin \Sigma$ as follows:
%
%
\begin{equation}
	\label{eq:sharp-encoding}
	w_{1}^{l}\# w_{2}^{l} \# ... \# w_{n}^{l}= w_{1}^{r}\# w_{2}^{r} \# ... \# w_{n}^{r},
\end{equation}

Then, we construct the graph representations for the  word equation in \eqref{eq:sharp-encoding}.
A graph representation $G=(V,E,V_{\text{T}},V_{\text{Var}})$ of a word equation consists of a set of nodes $V$, a set of edges $E \subseteq V \times V$, a set of terminal nodes $V_{\text{T}}\subseteq V$, and a set of variable nodes $V_{\text{Var}} \subseteq V$.
We start constructing the graph by drawing the ``$=$'' symbol as the root node. Its left and right children are the leftmost terms of both sides of the equation, respectively.
The rest of the graph is built following the choice of the graph type:

\begin{itemize}
\item \textbf{Graph 1:} Inspired by Abstract Syntax Trees (ASTs). Each letter and variable is represented by its own node, and words are represented by singly-linked lists of nodes.
    
    \item \textbf{Graph 2:} An extension of Graph 1, introducing additional edges from each term node back to the root node.
    
    \item \textbf{Graph 3:} An extension of Graph 1 which incorporates unique variable nodes. In this design, nodes representing variables are added, which are connected to their respective occurrences in the linked lists.
    This representation aims at facilitating the learning of long-distance variable relationships by GNNs.
    
    \item \textbf{Graph 4:} Similar in approach to Graph 3, but introducing unique nodes for letters instead of variables.
    
    \item \textbf{Graph 5:} A composite structure that merges the concepts of Graphs 3 and 4. It includes unique nodes for both variables and letters..
\end{itemize}

\begin{figure}[t]
  \begin{center}
		\includegraphics[width=\textwidth,trim=20 15 25 10]{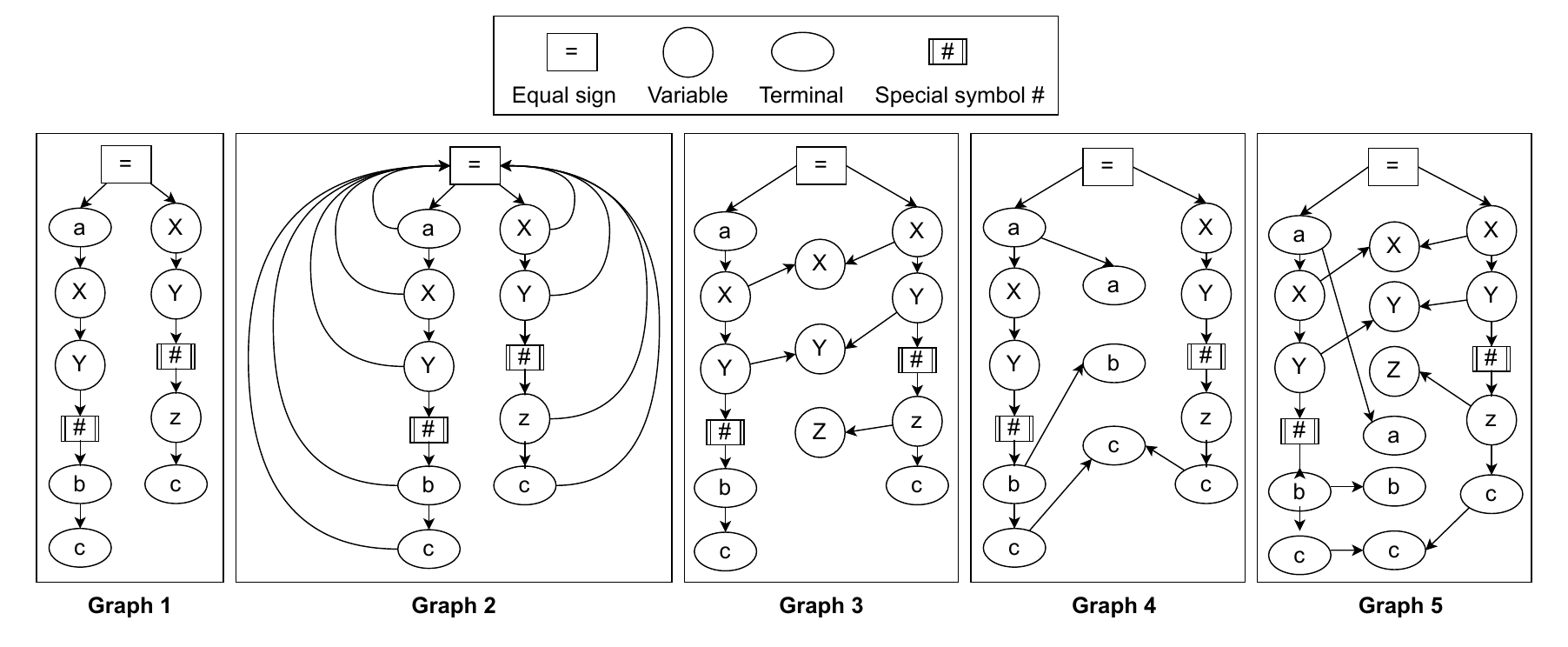}
  \end{center}

  \caption{The five graph representations for the word equation $aXY\# bc= XY\# Zc$ where $X, Y, Z$ are variables and $a,b, c$ are terminals
  }\label{fig:word-equation-graphs}
\end{figure}

Figure~\ref{fig:word-equation-graphs} illustrates the five graph representations of the conjunction of word equation $aXY\# bc= XY\# Zc$, where $\{X,Y,Z\}\subseteq \Gamma$ and $\{a,b,c\}\subseteq \Sigma$.

\subsection{Training of Graph Neural Networks}
\subsubsection{Forward Propagation.}
In the \emph{orderBranches} function of Algorithm~\ref{algorithm:SolveEqs}, we sort the branches by using the predictions from a trained GNN model. This GNN model performs a multi-classification task. 
Given a list of branches~$(b_1, \dots, b_n)$ resulting from a rule application, we expect the trained GNN model to output a list of floating-point numbers~$\hat{Y}_{n}=(\hat{y}_{1}, \dots, \hat{y}_{n})$, representing priorities of the branches. 
A higher value for $\hat{y}_i$ indicates a higher priority of the branch. For instance, given a node with two children $b_1$ and $b_2$, the output from the model could be $\hat{Y_{2}} = (0.3, 0.7)$, expressing the prediction that $b_2$ will lead to a solution more quickly than $b_1$ and should be explored first.
We detail the process of deriving $\hat{Y}_{n}$ at each split point using GNNs, exemplified by using $n=2$. 


\subsubsection{Propagation on Graphs.}

To explain forward propagation, suppose a node labelled with
formula~$\phi_0$ in the proof is rewritten by applying rule~$R_7$,
resulting in direct children labelled with $\phi_1$, $\phi_2$. The
situation is similar for applications of $R_8$.

Formulas~$\phi_{0}, \phi_{1}$, and $\phi_{2}$ are transformed to graphs $G_{0}=(V^{0},E^{0},V_{\text{T}}^{0},V_{\text{Var}}^{0})$, $G_{1}=(V^{1},E^{1},V_{\text{T}}^{1},V_{\text{Var}}^{1})$, and $G_{2}=(V^{2},E^{2},V_{\text{T}}^{2},V_{\text{Var}}^{2})$, respectively, according to one
of the encodings in Section~\ref{subsection:graph-representation}.
Each node in those graphs is then assigned an initial node
representation in $\mathbb{R}^m$, which is determined by the node
type: variable, letter, $=$, or $\#$. This gives rise to three initial
node representation functions~$H_i^0 : V^i \to \mathbb{R}^m$ for
$i \in \{1, 2, 3\}$, mapping the nodes of the graphs to vectors of real numbers.

Equation~\eqref{eq:GCN} defines how node representations are updated. Iterating the update rule, we obtain node representations~$H_i^{t} = \text{GCN}(H_i^{t-1}, E^i)$ for $i \in \{1, 2, 3\}$ and $t \in \{1, \ldots, T\}$, where the relation~$E^i$ is used to identify neighbours.
Subsequently, representation of the graphs as a whole are derived by summing up the node representations at point~$T$, resulting in $H_{G_i}= \sum_{v \in V^i} H_i^{T}(v)$.
	
Finally, these graph representations are concatenated and fed to a
classifier $\text{MLP}:(\mathbb{R}^{m})^3\rightarrow \mathbb{R}^{2}$
to calculate scores
$\hat{Y}_{2} = \text{MLP}(H_{G_{0}} || H_{G_{1}} || H_{G_{2}})$, where
$||$ denotes concatenation of vectors. The whole process generalizes
in a straightforward way to branching points in the proof tree with
$n$ children.

\subsubsection{Backward Propagation.}
The trainable parameters of the model, as described above, are the
initial node representations for the four types of graph nodes and the
parameters of the GCNs.
Those trainable parameters are optimized together
by minimizing the categorical cross-entropy loss between the predicted label $\hat{y}_{i}\in\hat{Y}_{n}$ and the true label $y_{i}\in Y_{n}$, using the following equation:

\begin{equation}
  \mathit{loss}=-\frac{1}{N}\sum_{1}^{N} y_{i} \log (\hat{y}_{i})	
\end{equation}
where $N$ is the number of split points in a training batch.
We explain how to collect the training data~$Y_{n}$ in the next section.

\subsection{Training Data Collection}

With our current algorithm, \UNSAT problems always require an exhaustive exploration of a proof tree; branch ordering therefore does not affect the solving time. We have thus focused on optimizing the process of finding solutions and only extract training data from \SAT problems. 

To collect our training labels, we construct the complete proof tree for given conjunctions of word equations, up to a certain depth. The tree enables us to identify cases of multiple \SAT pathways within the tree, and to identify situations where one branch leads to a solution more quickly than other branches.


Each node~$v$ of the proof tree with multiple children is labelled based on two criteria: the satisfiability status (\SAT, \UNSAT, or \UNKNOWN) of the formula, and the size of the proof sub-tree underneath each of the direct children.
Assume that node~$v$ has $n$ children, each of which has status \SAT, \UNSAT, or \UNKNOWN, respectively. If there is exactly one child of $v$, say the $i$'th child, that is \SAT, then the label of $v$ is a list of integers $(x_1, \ldots, x_n)$ with $x_i=1$ and $x_j = 0$ for $j \not= i$. If multiple children are \SAT, we examine the size of the sub-tree underneath each of those children, and label all  children with minimal sub-trees with $1$ in the list~$(x_1, \ldots, x_n)$.

More formally, suppose a proof tree $\tau = (N,\alpha,E,\lambda)$. The
satisfiability status~$\sigma(v)$ of a node~$v \in N$ is determined
by:
\begin{equation}
	\label{eq:split-point-satisfiability}
  \sigma(v) ~=~
  \begin{cases}
    \alpha(v) & \text{if~} \alpha(v) \in \{\SAT, \UNSAT, \UNKNOWN\}
    \\
    \SAT & \text{if there is~} u \in V \text{~with~} \sigma(u) = \SAT \text{~and~}
    (v, u) \in E
    \\
    \UNKNOWN &
    \begin{array}[t]{@{}l@{}}
      \text{otherwise, if there is~} u \in V \text{~with~} \sigma(u) = \UNKNOWN\\
      \quad \text{and~}    (v, u) \in E
    \end{array}
    \\
    \UNSAT & \text{otherwise}
  \end{cases}
\end{equation}
The size~$\Delta(v)$ of the sub-tree underneath a node~$v$ is defined by:
\begin{equation*}
  \Delta(v) ~=~ 1 + \sum_{u \in N, (v, u) \in E} \Delta(u)
\end{equation*}

Finally, the label $Y^v_{n}=(y_{1}, ..., y_{n})$ of a node~$v$ with $\sigma(v)=\text{\SAT}$ and
$n$ children~$v_1, \ldots, v_n$, where $y_{i}\in \{0,1\}$, is defined by:
\begin{align*}
  y_i &~=~
        \begin{cases}
          1 & \text{if~} \sigma(v_i) = \SAT \text{~and~} \Delta(v_i) = \min S
          \\
          0 & \text{otherwise}
        \end{cases}
  \\
  S &~=~ \{ \Delta(v_i) \mid \sigma(v_i) = \SAT \}
\end{align*}
When $\sum_{i=1}^{n}y_i >1$, we discard some children with label 1
until $\sum_{i=1}^{n}y_i =1$ to make sure that the label for each
split point is consistent.

\subsection{Guidance for the Split Algorithm using
the GNN Model}
\label{section:guide-split-algorithm}
In Algorithm~\ref{algorithm:SolveEqs}, we introduce five strategies
for the \textit{orderBranches} function implementation, designed to
evaluate the efficiency of deterministic versus stochastic methods in
branch ordering and to investigate the interplay between fixed and variable branch ordering approaches:
\begin{itemize}
	\item \textbf{Fixed Order:} Use a predetermined branch order, defined before execution. In our experiments, we simply use the order in which the branches are displayed in Figure~\ref{fig:split_rules}.
	\item \textbf{Random Order:} Reorder branches randomly.
	\item \textbf{GNN (S1):} Exclusively use the GNN model for branch ordering.
	\item \textbf{GNN-fixed (S2):} A balanced approach with a 50\% chance of using the GNN model and a 50\% chance of using the fixed order.
	\item \textbf{GNN-random (S3):} Similar to S2, but with the alternative 50\% chance dedicated to random ordering.
\end{itemize}


\section{Experimental Results}

This section presents the benchmarks used for our experiments and details the results with the different versions of our algorithm. It also provides a comprehensive comparison with other state-of-the-art solvers. 

\subsection{Implementation of DragonLi}
\OURSOLVER~\cite{RepositoryReference} is developed from scratch using \texttt{Python 3.8}~\cite{10.5555/1593511}. We train the models with \texttt{PyTorch}~\cite{NEURIPS2019_9015} and construct the GNNs using the \texttt{Deep Graph Library (DGL)}~\cite{wang2019dgl}. For tracking and visualizing training experiments, \texttt{mlflow}~\cite{10.1145/3399579.3399867} is employed. Proof trees and graph representations of word equations are stored in \texttt{JSON}~\cite{pezoa2016foundations} format, while 
\texttt{graphviz}~\cite{Ellson2004} is utilized for their tracking and visualization.

\subsection{Benchmarks Selection}
We consider two kinds of benchmarks: benchmarks that are artificially generated based on the benchmarks used to evaluate the solver \textsf{Woorpje}~\cite{day2019solving}, as well as benchmarks extracted from the non-incremental QF\_S, QF\_SLIA, and QF\_SNLIA track of the SMT-LIB benchmarks~\cite{smtlib:benchmark}. 
We summarize the benchmarks as following:

\begin{itemize}
\item \textbf{Benchmark 1} is generated by the mechanism used in \textsf{Woorpje} track~I.
Given finite sets of letters~$C$ and variables~$V$, we construct a string $s$ with maximum length of $k$ by randomly concatenating selected letters from \(C\).
We then form a word equation $s = s$ and repeatedly replace substrings in $s$ with the concatenation of between $1$ and $5$ fresh variables.
This procedure guarantees that the constructed word equation is \SAT.

\item \textbf{Benchmark 2} is generated by the mechanism used in \textsf{Woorpje} track~III.
It first generates a word equation using the following definition:
\begin{multline}
	\label{eq:track02}
	X_{n}aX_{n}bX_{n-1}bX_{n-2}\cdots bX_{1} =\\
        aX_{n}X_{n-1}X_{n-1}bX_{n-2}X_{n-2}b \cdots b X_{1}X_{1}baa
\end{multline}
where $X_{1}, ..., X_{n}$ are variables and $a$ and $b$ are letters.
We then generate a word equation using the  mechanism for Benchmark~1, and replace letters~$b$ in  \eqref{eq:track02} randomly with the left-hand side or the right-hand side of that equation.

\item \textbf{Benchmark 3} is generated by conjoining multiple word equations that were randomly generated using the mechanism described in Benchmark 1. This procedure mainly produces benchmarks that are \UNSAT.

\item \textbf{Benchmark 4} is extracted from benchmarks from the non-incremental QF\_S, QF\_SLIA, and QF\_SNLIA tracks of SMT-LIB. We obtain word equations by removing length constraints, regular expressions, and unsupported Boolean operators, which are not considered in this paper. As a result, benchmarks after transformation can be \SAT even if the original SMT-LIB benchmarks were \UNSAT.

\end{itemize}

\begin{table}[t]
	\caption{Number of SAT ($\checkmark$), UNSAT ($\times$), UNKNOWN ($\infty$), and evaluation (Eval) problems in the four benchmarks 
	}
	\label{table:benchmark}

        \begin{center}
        \begin{tabular}{|cccc|cccc|cccc|cccc|}
		\hline
          \multicolumn{4}{|c|}{Benchmark 1} & \multicolumn{4}{c|}{Benchmark 2} & \multicolumn{4}{c|}{Benchmark 3} & \multicolumn{4}{c|}{Benchmark 4} \\ 
          \multicolumn{4}{|c|}{Total: 3000} & \multicolumn{4}{c|}{Total: 21000} & \multicolumn{4}{c|}{Total: 41000} & \multicolumn{4}{c|}{Total: 2310} \\ \hline\hline
		\multicolumn{3}{|c|}{2000} & \multirow{2}{*}{Eval} & \multicolumn{3}{c|}{20000} & \multirow{2}{*}{Eval} & \multicolumn{3}{c|}{40000} & \multirow{2}{*}{Eval} & \multicolumn{3}{c|}{1855} & \multirow{2}{*}{Eval} \\ \cline{1-3} \cline{5-7} \cline{9-11} \cline{13-15} 
		\multicolumn{1}{|c|}{$\checkmark$} & \multicolumn{1}{c|}{$\times$} & \multicolumn{1}{c|}{$\infty$} &  & \multicolumn{1}{c|}{$\checkmark$} & \multicolumn{1}{c|}{$\times$} & \multicolumn{1}{c|}{$\infty$} &  & \multicolumn{1}{c|}{$\checkmark$} & \multicolumn{1}{c|}{$\times$} & \multicolumn{1}{c|}{$\infty$} &  & \multicolumn{1}{c|}{$\checkmark$} & \multicolumn{1}{c|}{$\times$} & \multicolumn{1}{c|}{$\infty$} &  \\ \hline
		\multicolumn{1}{|c|}{1997} & \multicolumn{1}{c|}{0} & \multicolumn{1}{c|}{3} & 1000 & \multicolumn{1}{c|}{1293} & \multicolumn{1}{c|}{0} & \multicolumn{1}{c|}{18707} & 1000 & \multicolumn{1}{c|}{1449} & \multicolumn{1}{c|}{1137} & \multicolumn{1}{c|}{37414} & 1000 & \multicolumn{1}{c|}{1673} & \multicolumn{1}{c|}{16} & \multicolumn{1}{c|}{166} & 455 \\ \hline
	\end{tabular}
        \end{center}

\end{table}

Table~\ref{table:benchmark} presents the number of problems in each benchmark. Benchmark 4 originates from a collection of 100805 SMT-LIB problems; after transformation,  we obtain 2310 problems. For evaluation, we selected hold-out sets of $1000$ (Benchmarks~1--3) and $455$ (Benchmark~4) problems were selected uniformly at random; those sets were exclusively used for evaluation, not for training or for tuning hyper-parameters.
All benchmarks, as well as our implementation and chosen hyper-parameters are available on Zenodo~\cite{ZenodoReference}.

We then applied the split algorithm
(Algorithm~\ref{algorithm:SolveEqs}) to all benchmarks with the \emph{fixed} reordering strategy, to determine 
the number of \SAT, \UNSAT and \UNKNOWN problems. 
After the dispatch phase, we only retained \SAT problems for the construction of the training dataset.



\begin{table}[]
		\caption{Evaluation on three metrics for different solvers. The labels SAT, UNS, UNI, CS, and CU are abbreviation of SAT, UNSAT, unique solved, commonly solved SAT, and commonly solved UNSAT, respectively.  The labels Fixed, Random, and GNN are the three variations of \OURSOLVER in Section~\ref{section:guide-split-algorithm}. Entries marked ``-" do not apply. Values less than 0.1 are rounded to 0.1. GNN rows for benchmarks 1-4 share the configuration ($BT_{2},S1,G5$).
		}\label{tbl:evaluation-table-1}
                \begin{center}
\small                  
\begin{tabular}{|c|c|ccccc|cccc|}
	\hline
	\multirow{2}{*}{Bench} & \multirow{2}{*}{Solver} & \multicolumn{5}{c|}{\begin{tabular}[c]{@{}c@{}}Number of solved\\ problems\end{tabular}} & \multicolumn{4}{c|}{\begin{tabular}[c]{@{}c@{}}Average \\ solving time (split numbet)\end{tabular}} \\ \cline{3-11} 
	&  & \multicolumn{1}{c|}{SAT} & \multicolumn{1}{c|}{UNS} & \multicolumn{1}{c|}{UNI} & \multicolumn{1}{c|}{CS} & CU & \multicolumn{1}{c|}{SAT} & \multicolumn{1}{c|}{UNS} & \multicolumn{1}{c|}{CS} & CU \\ \hline
	\multirow{8}{*}{\begin{tabular}[c]{@{}c@{}}1\\ (1000\\ SAT)\end{tabular}} & Fixed & \multicolumn{1}{c|}{999} & \multicolumn{1}{c|}{-} & \multicolumn{1}{c|}{0} & \multicolumn{1}{c|}{\multirow{8}{*}{777}} & \multirow{8}{*}{0} & \multicolumn{1}{c|}{\begin{tabular}[c]{@{}c@{}}4.1\\ (182.0)\end{tabular}} & \multicolumn{1}{c|}{- (-)} & \multicolumn{1}{c|}{\begin{tabular}[c]{@{}c@{}}4.0\\ (169.0)\end{tabular}} & - (-) \\ \cline{2-5} \cline{8-11} 
	& Random & \multicolumn{1}{c|}{996} & \multicolumn{1}{c|}{-} & \multicolumn{1}{c|}{0} & \multicolumn{1}{c|}{} &  & \multicolumn{1}{c|}{\begin{tabular}[c]{@{}c@{}}4.2\\ (349.6)\end{tabular}} & \multicolumn{1}{c|}{- (-)} & \multicolumn{1}{c|}{\begin{tabular}[c]{@{}c@{}}4.1\\ (269.8)\end{tabular}} & - (-) \\ \cline{2-5} \cline{8-11} 
	& GNN & \multicolumn{1}{c|}{995} & \multicolumn{1}{c|}{-} & \multicolumn{1}{c|}{0} & \multicolumn{1}{c|}{} &  & \multicolumn{1}{c|}{\begin{tabular}[c]{@{}c@{}}7.6\\ (215.7)\end{tabular}} & \multicolumn{1}{c|}{- (-)} & \multicolumn{1}{c|}{\begin{tabular}[c]{@{}c@{}}7.0\\ (162.3)\end{tabular}} & - (-) \\ \cline{2-5} \cline{8-11} 
	& \textsf{cvc5} & \multicolumn{1}{c|}{\textbf{1000}} & \multicolumn{1}{c|}{-} & \multicolumn{1}{c|}{0} & \multicolumn{1}{c|}{} &  & \multicolumn{1}{c|}{0.1 (-)} & \multicolumn{1}{c|}{- (-)} & \multicolumn{1}{c|}{0.1 (-)} & - (-) \\ \cline{2-5} \cline{8-11} 
	& \textsf{Ostrich} & \multicolumn{1}{c|}{918} & \multicolumn{1}{c|}{-} & \multicolumn{1}{c|}{0} & \multicolumn{1}{c|}{} &  & \multicolumn{1}{c|}{20.4 (-)} & \multicolumn{1}{c|}{- (-)} & \multicolumn{1}{c|}{19.6 (-)} & - (-) \\ \cline{2-5} \cline{8-11} 
	& \textsf{Woorpje} & \multicolumn{1}{c|}{967} & \multicolumn{1}{c|}{-} & \multicolumn{1}{c|}{0} & \multicolumn{1}{c|}{} &  & \multicolumn{1}{c|}{1.6 (-)} & \multicolumn{1}{c|}{- (-)} & \multicolumn{1}{c|}{0.5 (-)} & - (-) \\ \cline{2-5} \cline{8-11} 
	& \textsf{Z3} & \multicolumn{1}{c|}{902} & \multicolumn{1}{c|}{-} & \multicolumn{1}{c|}{0} & \multicolumn{1}{c|}{} &  & \multicolumn{1}{c|}{3.4 (-)} & \multicolumn{1}{c|}{- (-)} & \multicolumn{1}{c|}{2.4 (-)} & - (-) \\ \cline{2-5} \cline{8-11} 
	& \textsf{Z3-Noodler} & \multicolumn{1}{c|}{935} & \multicolumn{1}{c|}{-} & \multicolumn{1}{c|}{0} & \multicolumn{1}{c|}{} &  & \multicolumn{1}{c|}{1.9 (-)} & \multicolumn{1}{c|}{- (-)} & \multicolumn{1}{c|}{1.1 (-)} & - (-) \\ \hline
	\multirow{8}{*}{\begin{tabular}[c]{@{}c@{}}2\\ (1000\\ in\\ total)\end{tabular}} & Fixed & \multicolumn{1}{c|}{33} & \multicolumn{1}{c|}{0} & \multicolumn{1}{c|}{10} & \multicolumn{1}{c|}{\multirow{8}{*}{1}} & \multirow{8}{*}{0} & \multicolumn{1}{c|}{\begin{tabular}[c]{@{}c@{}}13.2\\ (1115.2)\end{tabular}} & \multicolumn{1}{c|}{- (-)} & \multicolumn{1}{c|}{4.8 (7)} & - (-) \\ \cline{2-5} \cline{8-11} 
	& Random & \multicolumn{1}{c|}{41} & \multicolumn{1}{c|}{0} & \multicolumn{1}{c|}{6} & \multicolumn{1}{c|}{} &  & \multicolumn{1}{c|}{\begin{tabular}[c]{@{}c@{}}11.7\\ (3879.5)\end{tabular}} & \multicolumn{1}{c|}{- (-)} & \multicolumn{1}{c|}{4.2 (60)} & - (-) \\ \cline{2-5} \cline{8-11} 
	& GNN & \multicolumn{1}{c|}{\textbf{71}} & \multicolumn{1}{c|}{0} & \multicolumn{1}{c|}{27} & \multicolumn{1}{c|}{} &  & \multicolumn{1}{c|}{\begin{tabular}[c]{@{}c@{}}46.0\\ (1813.5)\end{tabular}} & \multicolumn{1}{c|}{- (-)} & \multicolumn{1}{c|}{5.1 (5)} & - (-) \\ \cline{2-5} \cline{8-11} 
	& \textsf{cvc5} & \multicolumn{1}{c|}{4} & \multicolumn{1}{c|}{2} & \multicolumn{1}{c|}{4} & \multicolumn{1}{c|}{} &  & \multicolumn{1}{c|}{2.0 (-)} & \multicolumn{1}{c|}{0.1 (-)} & \multicolumn{1}{c|}{0.1 (-)} & - (-) \\ \cline{2-5} \cline{8-11} 
	& \textsf{Ostrich} & \multicolumn{1}{c|}{14} & \multicolumn{1}{c|}{\textbf{43}} & \multicolumn{1}{c|}{44} & \multicolumn{1}{c|}{} &  & \multicolumn{1}{c|}{40.7 (-)} & \multicolumn{1}{c|}{31.8 (-)} & \multicolumn{1}{c|}{2.5 (-)} & - (-) \\ \cline{2-5} \cline{8-11} 
	& \textsf{Woorpje} & \multicolumn{1}{c|}{23} & \multicolumn{1}{c|}{0} & \multicolumn{1}{c|}{2} & \multicolumn{1}{c|}{} &  & \multicolumn{1}{c|}{38.3 (-)} & \multicolumn{1}{c|}{- (-)} & \multicolumn{1}{c|}{0.1 (-)} & - (-) \\ \cline{2-5} \cline{8-11} 
	& \textsf{Z3} & \multicolumn{1}{c|}{6} & \multicolumn{1}{c|}{0} & \multicolumn{1}{c|}{2} & \multicolumn{1}{c|}{} &  & \multicolumn{1}{c|}{0.1 (-)} & \multicolumn{1}{c|}{- (-)} & \multicolumn{1}{c|}{4.2 (-)} & - (-) \\ \cline{2-5} \cline{8-11} 
	& \textsf{Z3-Noodler} & \multicolumn{1}{c|}{19} & \multicolumn{1}{c|}{0} & \multicolumn{1}{c|}{0} & \multicolumn{1}{c|}{} &  & \multicolumn{1}{c|}{45.8 (-)} & \multicolumn{1}{c|}{- (-)} & \multicolumn{1}{c|}{4.2 (-)} & - (-) \\ \hline
	\multirow{8}{*}{\begin{tabular}[c]{@{}c@{}}3\\ (1000\\ in \\ total)\end{tabular}} & Fixed & \multicolumn{1}{c|}{32} & \multicolumn{1}{c|}{79} & \multicolumn{1}{c|}{0} & \multicolumn{1}{c|}{\multirow{8}{*}{23}} & \multirow{8}{*}{50} & \multicolumn{1}{c|}{\begin{tabular}[c]{@{}c@{}}5.2\\ (1946.2)\end{tabular}} & \multicolumn{1}{c|}{\begin{tabular}[c]{@{}c@{}}65.8\\ (4227.0)\end{tabular}} & \multicolumn{1}{c|}{\begin{tabular}[c]{@{}c@{}}3.6\\ (57.0)\end{tabular}} & \begin{tabular}[c]{@{}c@{}}38.3\\ (796.6)\end{tabular} \\ \cline{2-5} \cline{8-11} 
	& Random & \multicolumn{1}{c|}{32} & \multicolumn{1}{c|}{79} & \multicolumn{1}{c|}{0} & \multicolumn{1}{c|}{} &  & \multicolumn{1}{c|}{\begin{tabular}[c]{@{}c@{}}9.5\\ (3861.8)\end{tabular}} & \multicolumn{1}{c|}{\begin{tabular}[c]{@{}c@{}}65.0\\ (4227.0)\end{tabular}} & \multicolumn{1}{c|}{\begin{tabular}[c]{@{}c@{}}3.8\\ (61.7)\end{tabular}} & \begin{tabular}[c]{@{}c@{}}38.5\\ (796.6)\end{tabular} \\ \cline{2-5} \cline{8-11} 
	& GNN & \multicolumn{1}{c|}{32} & \multicolumn{1}{c|}{65} & \multicolumn{1}{c|}{0} & \multicolumn{1}{c|}{} &  & \multicolumn{1}{c|}{\begin{tabular}[c]{@{}c@{}}214.3\\ (1471.2)\end{tabular}} & \multicolumn{1}{c|}{\begin{tabular}[c]{@{}c@{}}1471.2\\ (1471.2)\end{tabular}} & \multicolumn{1}{c|}{\begin{tabular}[c]{@{}c@{}}4.6\\ (63.7)\end{tabular}} & \begin{tabular}[c]{@{}c@{}}84.0\\ (796.6)\end{tabular} \\ \cline{2-5} \cline{8-11} 
	& \textsf{cvc5} & \multicolumn{1}{c|}{32} & \multicolumn{1}{c|}{943} & \multicolumn{1}{c|}{2} & \multicolumn{1}{c|}{} &  & \multicolumn{1}{c|}{0.1 (-)} & \multicolumn{1}{c|}{0.3 (-)} & \multicolumn{1}{c|}{0.1 (-)} & 0.3 (-) \\ \cline{2-5} \cline{8-11} 
	& \textsf{Ostrich} & \multicolumn{1}{c|}{27} & \multicolumn{1}{c|}{926} & \multicolumn{1}{c|}{0} & \multicolumn{1}{c|}{} &  & \multicolumn{1}{c|}{5.8 (-)} & \multicolumn{1}{c|}{4.7 (-)} & \multicolumn{1}{c|}{4.6 (-)} & 4.5 (-) \\ \cline{2-5} \cline{8-11} 
	& \textsf{Woorpje} & \multicolumn{1}{c|}{\textbf{34}} & \multicolumn{1}{c|}{723} & \multicolumn{1}{c|}{1} & \multicolumn{1}{c|}{} &  & \multicolumn{1}{c|}{12.4 (-)} & \multicolumn{1}{c|}{12.3 (-)} & \multicolumn{1}{c|}{0.1 (-)} & 23.2 (-) \\ \cline{2-5} \cline{8-11} 
	& \textsf{Z3} & \multicolumn{1}{c|}{26} & \multicolumn{1}{c|}{\textbf{953}} & \multicolumn{1}{c|}{10} & \multicolumn{1}{c|}{} &  & \multicolumn{1}{c|}{5.6 (-)} & \multicolumn{1}{c|}{0.5 (-)} & \multicolumn{1}{c|}{4.7 (-)} & 0.1 (-) \\ \cline{2-5} \cline{8-11} 
	& \textsf{Z3-Noodler} & \multicolumn{1}{c|}{28} & \multicolumn{1}{c|}{926} & \multicolumn{1}{c|}{0} & \multicolumn{1}{c|}{} &  & \multicolumn{1}{c|}{22.7 (-)} & \multicolumn{1}{c|}{0.3 (-)} & \multicolumn{1}{c|}{8.9 (-)} & 0.1 (-) \\ \hline
	\multirow{8}{*}{\begin{tabular}[c]{@{}c@{}}4\\ (455\\ in\\ total)\end{tabular}} & Fixed & \multicolumn{1}{c|}{416} & \multicolumn{1}{c|}{6} & \multicolumn{1}{c|}{0} & \multicolumn{1}{c|}{\multirow{8}{*}{403}} & \multirow{8}{*}{2} & \multicolumn{1}{c|}{\begin{tabular}[c]{@{}c@{}}5.1\\ (105.5)\end{tabular}} & \multicolumn{1}{c|}{\begin{tabular}[c]{@{}c@{}}17.7\\ (17119.5)\end{tabular}} & \multicolumn{1}{c|}{\begin{tabular}[c]{@{}c@{}}5.1\\ (51.0)\end{tabular}} & \begin{tabular}[c]{@{}c@{}}5.0\\ (246)\end{tabular} \\ \cline{2-5} \cline{8-11} 
	& Random & \multicolumn{1}{c|}{415} & \multicolumn{1}{c|}{6} & \multicolumn{1}{c|}{0} & \multicolumn{1}{c|}{} &  & \multicolumn{1}{c|}{\begin{tabular}[c]{@{}c@{}}4.9\\ (61.3)\end{tabular}} & \multicolumn{1}{c|}{\begin{tabular}[c]{@{}c@{}}17.9\\ (17119.5)\end{tabular}} & \multicolumn{1}{c|}{\begin{tabular}[c]{@{}c@{}}4.9\\ (38.1)\end{tabular}} & \begin{tabular}[c]{@{}c@{}}4.4\\ (246)\end{tabular} \\ \cline{2-5} \cline{8-11} 
	& GNN & \multicolumn{1}{c|}{418} & \multicolumn{1}{c|}{5} & \multicolumn{1}{c|}{0} & \multicolumn{1}{c|}{} &  & \multicolumn{1}{c|}{\begin{tabular}[c]{@{}c@{}}5.5\\ (118.3)\end{tabular}} & \multicolumn{1}{c|}{\begin{tabular}[c]{@{}c@{}}31.8\\ (5019.6)\end{tabular}} & \multicolumn{1}{c|}{\begin{tabular}[c]{@{}c@{}}5.3\\ (49.0)\end{tabular}} & \begin{tabular}[c]{@{}c@{}}8.8\\ (246)\end{tabular} \\ \cline{2-5} \cline{8-11} 
	& \textsf{cvc5} & \multicolumn{1}{c|}{406} & \multicolumn{1}{c|}{34} & \multicolumn{1}{c|}{0} & \multicolumn{1}{c|}{} &  & \multicolumn{1}{c|}{0.1 (-)} & \multicolumn{1}{c|}{0.1 (-)} & \multicolumn{1}{c|}{0.1 (-)} & 0.1 (-) \\ \cline{2-5} \cline{8-11} 
	& \textsf{Ostrich} & \multicolumn{1}{c|}{406} & \multicolumn{1}{c|}{6} & \multicolumn{1}{c|}{0} & \multicolumn{1}{c|}{} &  & \multicolumn{1}{c|}{1.4 (-)} & \multicolumn{1}{c|}{1.2 (-)} & \multicolumn{1}{c|}{1.4 (-)} & 1.2 (-) \\ \cline{2-5} \cline{8-11} 
	& \textsf{Woorpje} & \multicolumn{1}{c|}{\textbf{420}} & \multicolumn{1}{c|}{2} & \multicolumn{1}{c|}{0} & \multicolumn{1}{c|}{} &  & \multicolumn{1}{c|}{0.2 (-)} & \multicolumn{1}{c|}{3.6 (-)} & \multicolumn{1}{c|}{0.2 (-)} & 3.6 (-) \\ \cline{2-5} \cline{8-11} 
	& \textsf{Z3} & \multicolumn{1}{c|}{\textbf{420}} & \multicolumn{1}{c|}{10} & \multicolumn{1}{c|}{0} & \multicolumn{1}{c|}{} &  & \multicolumn{1}{c|}{0.1 (-)} & \multicolumn{1}{c|}{0.1 (-)} & \multicolumn{1}{c|}{0.1 (-)} & 0.1 (-) \\ \cline{2-5} \cline{8-11} 
	& \textsf{Z3-Noodler} & \multicolumn{1}{c|}{\textbf{420}} & \multicolumn{1}{c|}{\textbf{35}} & \multicolumn{1}{c|}{1} & \multicolumn{1}{c|}{} &  & \multicolumn{1}{c|}{0.1 (-)} & \multicolumn{1}{c|}{0.1 (-)} & \multicolumn{1}{c|}{0.1 (-)} & 0.1 (-) \\ \hline
\end{tabular}
                \end{center}
\end{table}


\begin{table}[t]
	\caption{Detailed results for number of solved problems of \OURSOLVER in terms of five graph representations (G1 to G5 represent Graph 1 to 5 in Section~\ref{subsection:graph-representation}), three strategies to apply GNN back to the algorithm (S1, S2, S3 in Section~\ref{section:guide-split-algorithm}), and three backtracking strategies ($BT_{1}$, $BT_{2}$, $BT_{3}$ in Section~\ref{section:GNN-Guided_Split_Algorithm}). The column GT denotes graph type.}
	\label{tbl:evaluation-table-2}
        \begin{center}          
\begin{tabular}{|c|ccccccccc|ccccccccc|}
	\hline
	\multirow{3}{*}{GT} & \multicolumn{9}{c|}{Benchmark 1} & \multicolumn{9}{c|}{Benchmark 2} \\ \cline{2-19} 
	& \multicolumn{3}{c|}{$BT_{1}$} & \multicolumn{3}{c|}{$BT_{2}$} & \multicolumn{3}{c|}{$BT_{3}$} & \multicolumn{3}{c|}{$BT_{1}$} & \multicolumn{3}{c|}{$BT_{2}$} & \multicolumn{3}{c|}{$BT_{3}$} \\ \cline{2-19} 
	& \multicolumn{1}{c|}{S1} & \multicolumn{1}{c|}{S2} & \multicolumn{1}{c|}{S3} & \multicolumn{1}{c|}{S1} & \multicolumn{1}{c|}{S2} & \multicolumn{1}{c|}{S3} & \multicolumn{1}{c|}{S1} & \multicolumn{1}{c|}{S2} & S3 & \multicolumn{1}{c|}{S1} & \multicolumn{1}{c|}{S2} & \multicolumn{1}{c|}{S3} & \multicolumn{1}{c|}{S1} & \multicolumn{1}{c|}{S2} & \multicolumn{1}{c|}{S3} & \multicolumn{1}{c|}{S1} & \multicolumn{1}{c|}{S2} & S3 \\ \hline
	G1 & \multicolumn{1}{c|}{991} & \multicolumn{1}{c|}{\textbf{1000}} & \multicolumn{1}{c|}{996} & \multicolumn{1}{c|}{997} & \multicolumn{1}{c|}{999} & \multicolumn{1}{c|}{999} & \multicolumn{1}{c|}{584} & \multicolumn{1}{c|}{627} & 624 & \multicolumn{1}{c|}{57} & \multicolumn{1}{c|}{48} & \multicolumn{1}{c|}{44} & \multicolumn{1}{c|}{53} & \multicolumn{1}{c|}{45} & \multicolumn{1}{c|}{40} & \multicolumn{1}{c|}{3} & \multicolumn{1}{c|}{3} & 3 \\ \hline
	G2 & \multicolumn{1}{c|}{998} & \multicolumn{1}{c|}{1000} & \multicolumn{1}{c|}{1000} & \multicolumn{1}{c|}{998} & \multicolumn{1}{c|}{1000} & \multicolumn{1}{c|}{998} & \multicolumn{1}{c|}{584} & \multicolumn{1}{c|}{627} & 624 & \multicolumn{1}{c|}{49} & \multicolumn{1}{c|}{46} & \multicolumn{1}{c|}{41} & \multicolumn{1}{c|}{53} & \multicolumn{1}{c|}{40} & \multicolumn{1}{c|}{50} & \multicolumn{1}{c|}{3} & \multicolumn{1}{c|}{3} & 3 \\ \hline
	G3 & \multicolumn{1}{c|}{997} & \multicolumn{1}{c|}{1000} & \multicolumn{1}{c|}{998} & \multicolumn{1}{c|}{998} & \multicolumn{1}{c|}{1000} & \multicolumn{1}{c|}{999} & \multicolumn{1}{c|}{584} & \multicolumn{1}{c|}{627} & 624 & \multicolumn{1}{c|}{53} & \multicolumn{1}{c|}{43} & \multicolumn{1}{c|}{49} & \multicolumn{1}{c|}{65} & \multicolumn{1}{c|}{46} & \multicolumn{1}{c|}{55} & \multicolumn{1}{c|}{3} & \multicolumn{1}{c|}{3} & 3 \\ \hline
	G4 & \multicolumn{1}{c|}{985} & \multicolumn{1}{c|}{999} & \multicolumn{1}{c|}{997} & \multicolumn{1}{c|}{984} & \multicolumn{1}{c|}{999} & \multicolumn{1}{c|}{995} & \multicolumn{1}{c|}{584} & \multicolumn{1}{c|}{627} & 624 & \multicolumn{1}{c|}{65} & \multicolumn{1}{c|}{52} & \multicolumn{1}{c|}{38} & \multicolumn{1}{c|}{59} & \multicolumn{1}{c|}{54} & \multicolumn{1}{c|}{44} & \multicolumn{1}{c|}{3} & \multicolumn{1}{c|}{3} & 3 \\ \hline
	G5 & \multicolumn{1}{c|}{995} & \multicolumn{1}{c|}{1000} & \multicolumn{1}{c|}{999} & \multicolumn{1}{c|}{995} & \multicolumn{1}{c|}{1000} & \multicolumn{1}{c|}{995} & \multicolumn{1}{c|}{584} & \multicolumn{1}{c|}{627} & 624 & \multicolumn{1}{c|}{64} & \multicolumn{1}{c|}{46} & \multicolumn{1}{c|}{44} & \multicolumn{1}{c|}{\textbf{71}} & \multicolumn{1}{c|}{54} & \multicolumn{1}{c|}{50} & \multicolumn{1}{c|}{3} & \multicolumn{1}{c|}{3} & 3 \\ \hline
	\multirow{3}{*}{GT} & \multicolumn{9}{c|}{Benchmark 3} & \multicolumn{9}{c|}{Benchmark 4} \\ \cline{2-19} 
	& \multicolumn{3}{c|}{$BT_{1}$} & \multicolumn{3}{c|}{$BT_{2}$} & \multicolumn{3}{c|}{$BT_{3}$} & \multicolumn{3}{c|}{$BT_{1}$} & \multicolumn{3}{c|}{$BT_{2}$} & \multicolumn{3}{c|}{$BT_{3}$} \\ \cline{2-19} 
	& \multicolumn{1}{c|}{S1} & \multicolumn{1}{c|}{S2} & \multicolumn{1}{c|}{S3} & \multicolumn{1}{c|}{S1} & \multicolumn{1}{c|}{S2} & \multicolumn{1}{c|}{S3} & \multicolumn{1}{c|}{S1} & \multicolumn{1}{c|}{S2} & S3 & \multicolumn{1}{c|}{S1} & \multicolumn{1}{c|}{S2} & \multicolumn{1}{c|}{S3} & \multicolumn{1}{c|}{S1} & \multicolumn{1}{c|}{S2} & \multicolumn{1}{c|}{S3} & \multicolumn{1}{c|}{S1} & \multicolumn{1}{c|}{S2} & S3 \\ \hline
	G1 & \multicolumn{1}{c|}{32} & \multicolumn{1}{c|}{32} & \multicolumn{1}{c|}{31} & \multicolumn{1}{c|}{32} & \multicolumn{1}{c|}{32} & \multicolumn{1}{c|}{32} & \multicolumn{1}{c|}{14} & \multicolumn{1}{c|}{16} & 16 & \multicolumn{1}{c|}{413} & \multicolumn{1}{c|}{415} & \multicolumn{1}{c|}{413} & \multicolumn{1}{c|}{417} & \multicolumn{1}{c|}{417} & \multicolumn{1}{c|}{416} & \multicolumn{1}{c|}{400} & \multicolumn{1}{c|}{404} & 402 \\ \hline
	G2 & \multicolumn{1}{c|}{34} & \multicolumn{1}{c|}{32} & \multicolumn{1}{c|}{32} & \multicolumn{1}{c|}{34} & \multicolumn{1}{c|}{32} & \multicolumn{1}{c|}{34} & \multicolumn{1}{c|}{13} & \multicolumn{1}{c|}{16} & 16 & \multicolumn{1}{c|}{416} & \multicolumn{1}{c|}{416} & \multicolumn{1}{c|}{416} & \multicolumn{1}{c|}{418} & \multicolumn{1}{c|}{418} & \multicolumn{1}{c|}{417} & \multicolumn{1}{c|}{400} & \multicolumn{1}{c|}{402} & 403 \\ \hline
	G3 & \multicolumn{1}{c|}{32} & \multicolumn{1}{c|}{32} & \multicolumn{1}{c|}{32} & \multicolumn{1}{c|}{31} & \multicolumn{1}{c|}{32} & \multicolumn{1}{c|}{33} & \multicolumn{1}{c|}{13} & \multicolumn{1}{c|}{16} & 16 & \multicolumn{1}{c|}{416} & \multicolumn{1}{c|}{416} & \multicolumn{1}{c|}{417} & \multicolumn{1}{c|}{418} & \multicolumn{1}{c|}{417} & \multicolumn{1}{c|}{415} & \multicolumn{1}{c|}{400} & \multicolumn{1}{c|}{402} & 404 \\ \hline
	G4 & \multicolumn{1}{c|}{\textbf{35}} & \multicolumn{1}{c|}{32} & \multicolumn{1}{c|}{32} & \multicolumn{1}{c|}{34} & \multicolumn{1}{c|}{32} & \multicolumn{1}{c|}{32} & \multicolumn{1}{c|}{14} & \multicolumn{1}{c|}{16} & 16 & \multicolumn{1}{c|}{414} & \multicolumn{1}{c|}{413} & \multicolumn{1}{c|}{415} & \multicolumn{1}{c|}{417} & \multicolumn{1}{c|}{417} & \multicolumn{1}{c|}{416} & \multicolumn{1}{c|}{400} & \multicolumn{1}{c|}{403} & 403 \\ \hline
	G5 & \multicolumn{1}{c|}{31} & \multicolumn{1}{c|}{31} & \multicolumn{1}{c|}{31} & \multicolumn{1}{c|}{32} & \multicolumn{1}{c|}{31} & \multicolumn{1}{c|}{32} & \multicolumn{1}{c|}{14} & \multicolumn{1}{c|}{16} & 16 & \multicolumn{1}{c|}{415} & \multicolumn{1}{c|}{416} & \multicolumn{1}{c|}{414} & \multicolumn{1}{c|}{\textbf{418}} & \multicolumn{1}{c|}{417} & \multicolumn{1}{c|}{416} & \multicolumn{1}{c|}{400} & \multicolumn{1}{c|}{402} & 402 \\ \hline
\end{tabular}
        \end{center}
\end{table}


\subsection{Experimental Settings}

\OURSOLVER was parametrized with values $l_{BT2}=500$,  $l_{BT2}^{step}=250, l_{BT3}=20$ for Algorithm~\ref{algorithm:SolveEqs}. In addition, we chose a hidden layer of size 128 for all neural networks, and used a two message-passing layer (i.e. $t=2$ in Equation~\ref{eq:MPGNN}) for the GNN.
Each problem in the benchmarks is evaluated on a computer equipped with two Intel Xeon E5 2630 v4 at 2.20 GHz/core and 128GB memory. The GNNs are trained on A100 GPUs. 
We measured the number of solved problems and the average solving time (in seconds), with timeout of 300 seconds for each proof attempt.


\subsection{Comparison with Other Solvers}
In Table~\ref{tbl:evaluation-table-1}, we evaluate three versions of our algorithm based on their implementation of the \textit{orderRules} function: the fixed, random, and GNN-guided order versions (listed in Section~\ref{section:guide-split-algorithm}).
The performance of the GNN-guided \OURSOLVER (row GNN in Table~\ref{tbl:evaluation-table-1}) for each benchmark is selected from the best results out of 45 experiments (see Table~\ref{tbl:evaluation-table-2}).
These experiments use different combinations of five graph representations, three backtrack strategies, and three GNN guidance strategies, as shown in bold text in Table~\ref{tbl:evaluation-table-2}.
We compare the results with those of five other solvers: \textsf{Z3} (v4.12.2)~\cite{demoura2008z}, \textsf{Z3-Noodler} (v1.1.0)~\cite{10.1007/978-3-031-57246-3_2}, \textsf{cvc5} (v1.0.8)~\cite{10.1007/978-3-030-99524-9_24}, \textsf{Ostrich} (v1.3)~\cite{DBLP:journals/pacmpl/ChenHLRW19}, and \textsf{Woopje} (v0.2)~\cite{day2019solving}.

The primary metric is the number of solved problems. \OURSOLVER outperforms all other solvers on \SAT problems in benchmark 2.
Notably, the GNN-based \OURSOLVER solves the highest number of \SAT problems.
For the conjunction of multiple word equations (benchmark 3 and 4), \OURSOLVER's performance is comparable to the other solvers. The order of processing word equations is  crucial for those problems; currently, our solver uses only a predefined sequence, indicating significant potential for improvement.
A limitation of \OURSOLVER is determining \UNSAT cases, as it requires an exhaustive check of all nodes in the proof tree.

In terms of average solving time for solved problems, GNN-based \OURSOLVER does not hold an advantage.
This is mainly due to the overhead associated with encoding equations into a graph at each split point and invoking GNNs. Furthermore, \OURSOLVER is written in Python and not particularly optimized at this point, so that there is ample room for improvement in future efforts.

The measurement of the average number of splits in solved problems is used to gain insight into the efficiency of the different versions of our algorithm.
For benchmark 1 and 2, the GNN-guided version outperforms the others on the commonly solved problems.
However, for benchmarks 3 and 4, the GNN-guided version does not show advantages.
This is for the same reason mentioned in the metric of the number of solved problems; namely, the performance is also influenced by the order of processing equations when dealing with the conjunction of multiple word equations.

We summarize the experimental results compared with some of the leading string solvers as follows:
\begin{enumerate}
	\item \OURSOLVER shows better or comparable performance on SAT problems but is limited on UNSAT problems. This occurs because the split algorithm concludes SAT upon finding one SAT node, but can conclude UNSAT only after exhaustively exploring the proof tree. In contrast, other solvers may invest more time in proving UNSAT cases, for instance by reasoning about string length or Parikh vectors.
	\item \OURSOLVER performs better than other solvers on single word equations (e.g., benchmark 2) and comparably on conjunctions of multiple word equations.
	This performance difference is because the initial choice of word equation for splitting is crucial for the split algorithm, and this aspect is not optimized currently.
	\item Incorporating GNN guidance into the proof tree search enhances the performance of the pure split algorithm for \SAT problems, but currently does not lead to an improvement for \UNSAT problems.
\end{enumerate}

\subsection{Ablation Study}
Table~\ref{tbl:evaluation-table-2} displays the number of problems solved in 45 experiments across all benchmarks using the GNN-guided version.

In terms of backtrack strategies, $BT_1$ performs a pure depth-first search, but it already has good performance.
$BT_2$ performs a depth-first search controlled by parameters $l_{BT_{2}}$ and $l_{BT_{2}}^{step}$, and in many cases, it delivers the best performance.
$BT_3$ conducts a systematic search on the proof tree, which is complete for proving problems \SAT, but turns out to be relatively inefficient in the experiments and solves the fewest problems given a fixed timeout.  This indicates that more sophisticated search strategies may lead to even better performance.

In terms of the guiding strategies (S1, S2, S3), using the GNN alone (S1) to guide the branch order is better than combining it with predefined and random orders (S2 and S3) in most cases. This indicates that the GNN model successfully learns useful patterns at each split point and can be used as a stand-alone heuristic for branching.

In terms of the five graph representations, Graph 1 has the simplest structure, which represents the syntactic information of the word equations and thus incurs the least overhead when we call the model at each split point.
This yields average performance compared to other graph representations.
The performances of Graph 2 are weaker than others; this is probably due to the extra edges not providing any benefits for prediction, but leading to additional computational overhead.   
Graphs 3 and 4 emphasize the relationships between terminals and variables, respectively, thus the performance is biased by individual problems. 
Graph 5 considers the relationships for both terminals and variables, thus it has bigger overhead than Graphs 1, 3, and 4, but it offers relatively good performance. This shows that representing semantic information of the word equations well in graphs helps the model to learn important patterns.
In summary, the setting $(BT_{2}, S1, Graph5)$ performs the best.


\section{Related Work}

There are many techniques within solvers supporting word equations, as well as in stand-alone word equation solvers.
For instance, the SMT solvers \textsf{Norn}~\cite{10.1007/978-3-319-21690-4_29} and \textsf{TRAU}~\cite{8602997} introduce several improvements on the inference rules~\cite{10.1007/978-3-319-08867-9_10}, including length-guided splitting of equalities and a more efficient way to handle disequalities.
The stand-alone word equation solver \textsf{Woorpje}~\cite{day2019solving} reformulates the word equation problem as a reachability problem for nondeterministic finite automata, then encodes it as a propositional satisfiability problem which can be handled by SAT solvers.
In~\cite{10.1145/3372020.3391556}, the authors propose a transformation system that extends the Nielsen transformation~\cite{levi1944semigroups} to work with linear length constraints. 
This transformation system can be integrated into existing string solvers such as Z3STR3~\cite{8102241}, Z3SEQ~\cite{demoura2008z}, and CVC4~\cite{10.1007/978-3-642-22110-1_14}, thereby advancing the efficiency of word equation resolution.

GNNs excel at analyzing the graph-like structures of logic formulae, offering a complementary approach to formal verification. 
FormulaNet~\cite{10.5555/3294996.3295038} is an early study guiding the premise selection for Automated Theorem Provers (ATPs). It uses MP-GNNs~\cite{DBLP:journals/corr/GilmerSRVD17} to process the graph representation of the formulae in the proof trace extracted from HOL Light~\cite{10.1007/978-3-642-03359-9_4}. With more studies~\cite{10.1007/978-3-030-51054-1_29,DBLP:journals/corr/abs-1905-10006,LPAR2023:Guiding_an_Instantiation_Prover} exploring this path, this trend quickly expands to related fields.
For instance, for SAT solvers~\cite{wang2023neuroback,kurin2020improving}, NeuroSAT~\cite{DBLP:journals/corr/abs-1903-04671,selsam2019learning} predicts the probability of variables appearing in unsat cores to guide the variable branching decisions for Conflict-Driven Clause Learning (CDCL)~\cite{MarquesSilva1999GRASPAS}. 
Moreover, GNNs have been combined with various formal verification techniques, such as scheduling SMT \mbox{solvers}~\cite{9643296}, loop invariant reasoning~\cite{NEURIPS201865b1e92c,10.1007/978-3-030-53291-8-9}, or guiding Constraint Horn Clause (CHC) \mbox{solvers}~\cite{10.1007/978-3-031-50524-9_13,liang2022exploring,8603013}. They provide the empirical foundations for designing the learning task in Section~\ref{section:learning}, such as the graph representation of word equations and forming the learning task in split points.

\section{Conclusion and Future Work}
\label{section:conclusion-and-future-work}
This study introduces a GNN-guided split algorithm for solving word equations, along with five graph representations to enhance branch ordering through a multi-classification task at each split point of the proof tree.
We developed our solver from scratch instead of modifying a state-of-the-art SMT solver. This decision prevents the confounding influences of pre-existing optimizations in state-of-the-art SMT solvers, allowing us to isolate and evaluate the specific impact of GNN guidance more effectively.

We investigate various configurations, including graph representations, backtrack strategies, and the conditions for employing GNN-guided branches, aiming to analyze the behaviors of the algorithm across different settings.

The evaluation tables reveal that while the split algorithm effectively solves single word equations, it does not demonstrate marked improvements for multiple conjunctive word equations relative to other solvers. This discrepancy is attributed to the significance of the processing order for conjunctive word equations, where our current solver employs a predefined order. It is possible to make use of a GNN to compute the best equation to start with. However, this involves ranking a list of elements with variable lengths, rather than performing a fixed-category classification task, and requires completely different training for this specific task. Consequently, as future work, we aim to investigate both deterministic and stochastic strategies to optimize the ordering of conjunctive word equations for the split algorithm.
Our algorithm is also limited in handling ~~UNSAT problems because it can only conclude UNSAT by exhausting the proof tree. This can be improved in future work.

\paragraph{Acknowledgement.}
The computations were enabled by resources provided by the National Academic Infrastructure for Supercomputing in Sweden (NAISS) at Chalmers Centre for Computational Science and Engineering (C3SE) and Uppsala Multidisciplinary Center for Advanced Computational Science (UPPMAX) partially funded by the Swedish Research Council through grant agreement no. 2022-06725.
The research was also partially supported by the Swedish Research
Council through grant agreement no.~2021-06327, by a Microsoft Research PhD grant, the Swedish Foundation for Strategic Research (SSF) under the project WebSec (Ref.\ RIT17-0011), and the Wallenberg project UPDATE.

\clearpage

%
%
%
\bibliographystyle{splncs04}
\bibliography{mybibliography}

\clearpage

\appendix
\section{Proof of Lemma~\ref{lemma:1}}
\label{appx:proof-lemma-1}

\begin{proof}
	Outline of the proof:
	\begin{itemize}
		\item $R_{1}$: Trivial.
		\item $R_{2}$: Simplify $\epsilon = \epsilon \wedge \phi$ to $\phi$, then the premise and conclusion are identical.
		\item $R_{3}$: Assign $\epsilon$ to $X$ for both the premise and conclusion, then simplify it to make the premise and conclusion identical.
		\item $R_{4}$: Does not exist an assignment to make the premise SAT due to $a$ cannot be $\epsilon$.
		\item $R_{5}$, $R_{9}$: Simplify the premise to $u=v \wedge \phi$ by eliminating the same prefix, then the premise and conclusion are identical.
		\item $R_{6}$: Does not exist an assignment to make the premise SAT due to mismatched prefix. 
		\item $R_{7}$, $R_{8}$:  Assign $\epsilon$ or $a\cdot X'$ to $X$ to make the premise identical to one of the conclusions
		\item $R_{8}$: Assign $Y$ to $X$, $X$ to $Y$, $Y\cdot Y'$ to $X$,  or $X\cdot X'$ to $Y$ to make the premise identical to one of the conclusions.
	\end{itemize}
\end{proof}

\end{document}